\documentclass[UTF8,journal]{IEEEtran}
\usepackage{amsmath,amsfonts}
\usepackage{array}
\usepackage{epsfig}
\usepackage{graphicx,booktabs}
\usepackage{cite}
\usepackage{color}
\usepackage{amsthm}
\usepackage{threeparttable}
\usepackage{multirow}

\ifCLASSINFOpdf
\else
   \usepackage[dvips]{graphicx}
\fi
\usepackage{url}
\usepackage{graphicx}

\definecolor{gray}{rgb}{0.5,0.5,0.5}
\definecolor{popule}{RGB}{138,43,226}

\begin{document}

\title{Compressed Video Quality Enhancement with Temporal Group Alignment and Fusion}

\author{Qiang Zhu, 
	    Yajun Qiu,
	    Yu Liu,
        Shuyuan Zhu,~\IEEEmembership{Member, IEEE,}
        Bing Zeng,~\IEEEmembership{Fellow, IEEE}
 \thanks{The authors are with School of Information and Communication Engineering, University of Electronic Science and Technology of China, Chengdu 611731, China. This work was supported by the National Natural Science Foundation of China under Grant U20A20184, in part by the Natural Science Foundation of Sichuan Province under Grant 2023NSFSC1972,  2022NSFSC0509, and in part by CAAI-HuaweiMindSpore Open Fund under Grant CAAIXSJLJJ-2022-060A.}
 \vspace{-2.2em}
}

\maketitle

\begin{abstract}
In this paper, we propose a temporal group alignment and fusion network to enhance the quality of compressed videos by using the long-short term correlations between frames. The proposed model consists of the intra-group feature alignment (IntraGFA) module, the inter-group feature fusion (InterGFF) module, and the feature enhancement (FE) module. We form the group of pictures (GoP) by selecting frames from the video according to their temporal distances to the target enhanced frame. With this grouping, the composed GoP can contain either long- or short-term correlated information of neighboring frames. We design the IntraGFA module to align the features of frames of each GoP to eliminate the motion existing between frames. We construct the InterGFF module to fuse features belonging to different GoPs and finally enhance the fused features with the FE module to generate  high-quality video frames. The experimental results show that our proposed method achieves up to 0.05dB gain and lower complexity compared to the state-of-the-art method.
\end{abstract}

\begin{IEEEkeywords}
Compressed video, quality enhancement, long-short term, feature, correlation.
\end{IEEEkeywords}

\IEEEpeerreviewmaketitle
\section{Introduction}
\IEEEPARstart{C}{ompression} artifacts normally occur in the compressed video due to the lossy coding, sometimes resulting in serious quality degradation \cite{ref1,ref2,ref3}.
To address this issue, compressed video quality enhancement has been developed to construct high-quality (HQ) video from the compressed low-quality (LQ) one.

Over the past decades, lots of methods have been proposed to implement compressed video quality enhancement.
These methods achieve impressive quality improvement, not only for the 2D video {\cite{ref9-0, ref11,ref13}} but also for the 3D video \cite{ref38,ref39}.
Among these methods, the single-frame-based enhancement methods \cite{ref9-0,ref11,ref13} were firstly proposed and they utilized an LQ frame to generate HQ frame. Although these methods improved quality for compressed videos, they ignored using the temporal correlation of neighboring frames to construct the enhancement models, which limits their performance.

To achieve high enhancement performance, the multi-frame-based enhancement methods \cite{ref14,ref18,ref24,ref21-1,ref35,ref36,ref37} were proposed and the optical flow is often used in these methods to construct temporal correlation between frames for the producing of HQ videos. For instance, the multi-frame quality enhancement (MFQE) \cite{ref14} was firstly proposed to use convolution neural network (CNN) to estimate the optical flow of the local high-quality frame and the target frame. With the flow, the temporal correlation of frames was constructed in MFQE, which makes the model achieve high performance. In addition, a multi-scale motion compensation network was designed in MFQE 2.0 \cite{ref24} to utilize the multi-scale optical flow to generate the motion-compensated feature for removing compression artifacts. However, if there exist large motions in video, it is difficult to obtain accurate flow, which limits the enhancement efficiency. Recently, the spatio-temporal deformable fusion (STDF) \cite{ref18} was proposed for compressed video quality enhancement. It adopts the deformable convolution network (DCN) \cite{ref25} to explore temporal information than optical flow from the video and accordingly achieves significant performance improvement. Later on, the recursive fusion and deformable spatiotemporal attention (RFDA) \cite{ref21-1} based method \cite{ref18} was designed to exploit long-range spatio-temporal information and to pemploy the deformable spatiotemporal attention based on DCN for the construction of HQ videos. Recently, spatio-temporal detail information retrieval network \cite{ref35} was constructed to utilize DCN with different receptive fields to compose temporal detail information for quality enhancement.

In addition to the DCN-based methods, the attention-based methods were also proposed to improve the quality of compressed videos\cite{ref36} or the super-resolved videos \cite{ref41,ref42}. For instance, both spatio-temporal attention and channel attention were adopted in \cite{ref36} to exploit correlation of multiple frames for generation of HQ videos. The deformable attention was employed in \cite{ref41} to finely fuse features of adjacent frames for the generation of high-quality video. The temporal difference attention was proposed in \cite{ref42} based on temporal difference learning to explore motions from video and then to compensate them to target frame for the enhancement of video quality. However, the above methods disregard the utilization of the long and short terms correlations between frames, which limits their efficiency to obtain effective temporal features for the composition of high-quality videos.

In this paper, we propose a temporal group alignment and fusion network (TGAFNet) for the quality enhancement of compressed videos according to the long-short term correlations between frames. It is a post-processing method like \cite {ref24, ref18, ref21-1} and just carried out at decoder side. Our proposed model consists of the intra-group feature alignment (IntraGFA) module, the inter-group feature fusion (InterGFF) module and feature enhancement (FE) module. The temporal grouping strategy has demonstrated high feature alignment performance for quality improvement of super-resolved video\cite{ref40}. Inspired by this, we also adopt this strategy in our proposed method. Specifically, we firstly select the frames from video to form group of pictures (GoP) according to the temporal distances to the target enhanced frame. Then, the IntraGFA module is applied to align the features of frames of each GoP to eliminate the motion error. After that, the aligned features of different GoPs are fed into the InterGFF module to gradually fuse the aligned features to produce the temporally-fused features for the construction of high-quality video frame. Finally, the FE module which is constructed based on our proposed spatial dual contextual block is applied to enhance the features for the generation of detailed video contents.

\vspace{-0.8em}
\section{Proposed Method}

\begin{figure*}
	\vspace{-2.0em}
	\centerline{\includegraphics[width=16.0cm]{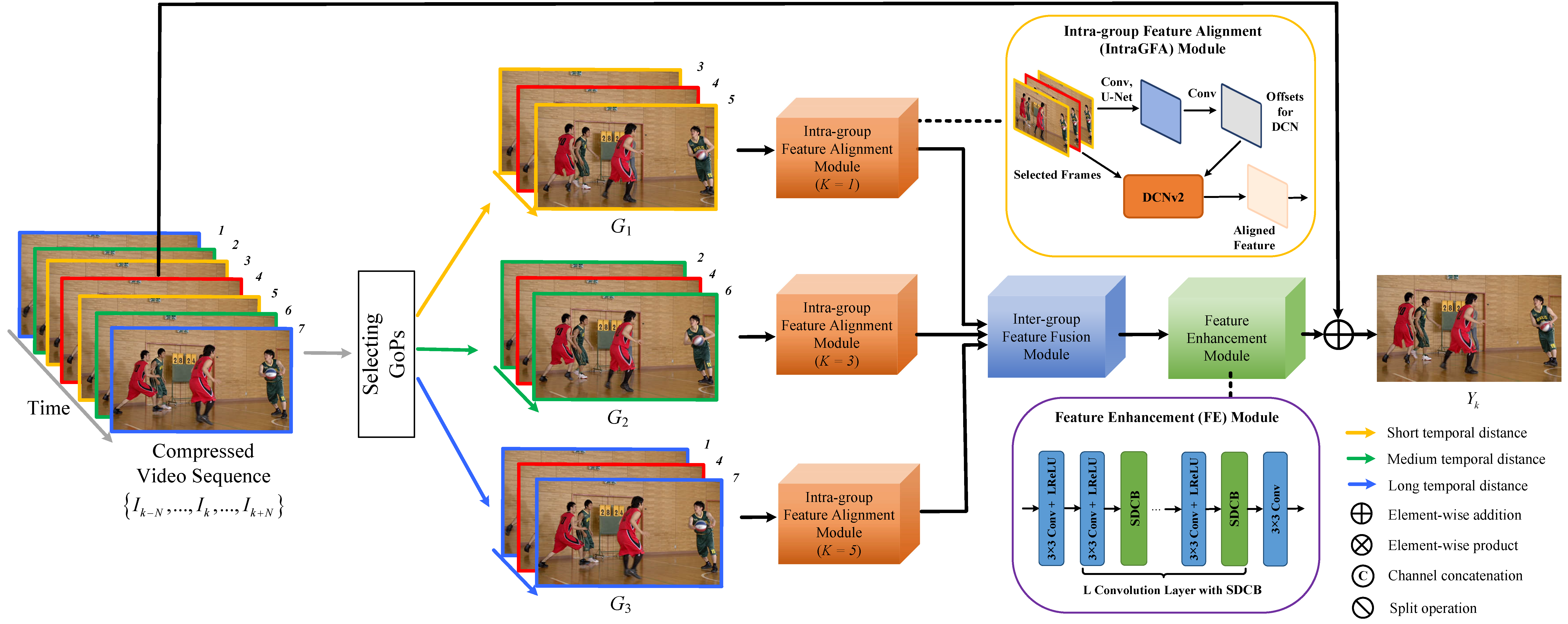}}
	\caption{ The proposed model which consists of the InterGFA, IntraGFF and FE modules. }
	\label{fig_pipline}
	\vspace{-1em}
\end{figure*}

The architecture of our proposed model, namely TGAFNet, is illustrated in Fig. \ref{fig_pipline} and it consists of the IntraGFA, InterGFF and FE modules. Given an LQ video $S$ =$\left\{\ldots, I_{k-N}, \ldots, I_{k-1}, I_k, I_{k+1}, \ldots, I_{k+N},\ldots \right\}$, we aim to produce the HQ frame $Y_k$ for the target frame $I_k$ according to the long-short term correlations between $I_k$ and its neighbors.

To achieve the above goal, we firstly form $N$ GoPs for each $I_k$ with $I_k$ and some of its neighbors, where GoP is formed as {$G_{i}$=$\left\{I_{k-i}, I_k, I_{k+i}\right\}$, $i=1,2, \ldots, N$}. The neighboring frames of $I_k$ in each $G_{i}$ are selected according to their distances to $I_k$. In this work, we specify $N$=3 and three groups $G_1$, $G_2$ and $G_3$ are formed for each $I_k$, as illustrated in Fig. \ref{fig_pipline}. In order to achieve the effective enhancement for $I_k$, we construct the IntraGFA module and InterGFF module to align and fuse features so that we can enhance the quality of compressed video by effectively using the long-short term correlations between frames. In addition, we design the FE module to enhance features to compose high-quality video frames.

In our work, the IntraGFA module is firstly applied to align the features of the frames belonging to each $G_{i}$. According to the temporal distances of neighboring frames to $I_k$, we design the IntraGFA module based on DCN \cite{ref25} with different kernel size $K$ to extract features. In this work, we specify that $K=1, 3$, and $5$ for $G_1$, $G_2$, and $G_3$, respectively. After obtaining features, we align them in each GoP to eliminate motion error within frames. Accordingly, the aligned features of each GoP are obtained as

\vspace{-1em}
\begin{equation}
\mathcal{F}^{(K)}=\operatorname{IntraGFA}(G_i).
\end{equation}

Then, we apply the InterGFF module to fuse the aligned features to obtain the temporally-fused feature as

\vspace{-.8em}
\begin{equation}
\mathcal{F}_f=\operatorname{InterGFF}(\mathcal{F}^{(1)},\mathcal{F}^{(3)},\mathcal{F}^{(5)}).
\end{equation}

After that, $\mathcal{F}_f$ is fed into the FE module to generate the enhanced feature

\vspace{-1em}
\begin{equation}
\mathcal{E}_k=\operatorname{FE}(\mathcal{F}_f).
\end{equation}

Finally, the HQ frame $Y_k$ is obtained by fusing $\mathcal{E}_k$ and the input frame $I_k$ as

\vspace{-1em}
\begin{equation}
Y_k=\mathcal{E}_k+I_k.
\end{equation}


\vspace{-1.2em}
\subsection{Intra-group Feature Alignment}  \label{R1} %

To implement the intra-group feature alignment, we construct the IntraGFA module based on U-Net \cite{ref035} and DCN \cite{ref25} to extract the features of frames of each GoP and align these features. Specifically, we firstly concatenate the frames of ${G_{i}}$ to form $f_{G_i}$ and apply U-Net, denoted as $\mathcal{U}_{net}$, to $f_{G_{i}}$ to extract features. Then, a convolution layer is performed on the features to predict the deformable offset $\Delta\mathcal{O}_{i}$ for DCN. The offset $\Delta\mathcal{O}_{i}$ is obtained by using the intra-group offset convolutional network $\operatorname{OCN_{intra}}$ as

\vspace{-.5em}
{\begin{equation}
\Delta\mathcal{O}_{i}=\operatorname{OCN_{intra}}(\mathcal{U}_{net}(f_{G_i}))  \in \mathbb{R}^{ H\times W \times 6K^2},
\end{equation}}
\vspace{-1em}

\noindent
where $\operatorname{OCN_{intra}}$ is composed by a convolution layer with $6K^2$ filters whose size is $3\times3$, $K$ is the kernel size of deformable convolution, and $H$ and $W$ are the height and width of feature, respectively. After that, the offset $\Delta\mathcal{O}_{i}$ is adopted in DCN to align $f_{G_{i}}$, generating the aligned features
\vspace{-0.3em}
\begin{equation}
\mathcal{F}^{(K)}=\operatorname{DCN}^{(K)}(f_{G_i}, \Delta\mathcal{O}_{i}) \in{\mathbb{R}}^{ H \times W \times C },
\end{equation}


\noindent
where $\operatorname{DCN}^{(K)}$ is the deformable convolution network whose kernel size is $K$, and $C$ is the channel number of feature. With the IntraGFA module, the aligned features of $\mathcal{F}^{(1)},\mathcal{F}^{(3)}$, $\mathcal{F}^{(5)}$ are generated for the corresponding three GoPs $G_1$, $G_2$, and $G_3$. Due to the temporal distance-based grouping strategy, the aligned features in each GoP contain the short-term or long-term correlations between frames.

\vspace{-1.0em}
\subsection{Inter-group Feature Fusion}  \label{R2}

To effectively utilize the long-short term correlations between frames to generate the effective features, we gradually fuse the aligned features to generate the temporally-fused feature. More specifically, we firstly fuse $\mathcal{F}^{(1)}$ with $\mathcal{F}^{(3)}$ and also fuse $\mathcal{F}^{(3)}$ with $\mathcal{F}^{(5)}$ to aggregate the temporal information of the neighboring groups to generate the fused features

\vspace{-0.4em}
\begin{equation}
	\mathcal{F}^{(1\mbox{-}3)}=\operatorname{Conv}([\mathcal{F}^{(1)},\mathcal{F}^{(3)}])
\end{equation}

\noindent
and
\vspace{-0.5em}
\begin{equation}
	\mathcal{F}^{(3\mbox{-}5)}=\operatorname{Conv}([\mathcal{F}^{(3)},\mathcal{F}^{(5)}]),
\end{equation}

\noindent
where $[\cdot]$ is the concatenation operation and $\operatorname{Conv}$ is a convolution layer with $C$ filters whose size is $3\times3$ and followed by a LeakyReLU activation function. Then, $\mathcal{F}^{(1\mbox{-}3)}$ and $\mathcal{F}^{(3\mbox{-}5)}$ are combined together to be sent into the residual channel attention block (RCAB) \cite{ref33-1}, forming the fused feature

\vspace{-0.5em}
\begin{equation}
\mathcal{F}_{fus}=\operatorname{RCAB}(\operatorname{Conv}[\mathcal{F}^{(1\mbox{-}3)},\mathcal{F}^{(3\mbox{-}5)}]).
\end{equation}

\noindent
Next, we use the inter-group offset convolution network $\operatorname{OCN_{inter}}$ and DCN to predict the deformable offset and align the fused feature to obtain the temporally-fused feature

\vspace{-0.8em}
\begin{equation}
	\mathcal{F}_{f}=\operatorname{DCN}^{(K)}(\mathcal{F}_{fus}, \operatorname{OCN_{inter}}(\mathcal{F}_{fus})) \in{\mathbb{R}}^{H \times W \times C},
\end{equation}

\noindent
where the kernel size of DCN is set as {$K=1$} and $\operatorname{OCN_{inter}}$ consists of a convolution layer with $2CK^2$ filters whose size is $3\times3$. With the InterGFF module, the fused feature is generated and it contains the long-term and short-term temporal information of the frames.

\vspace{-1.2em}
\subsection{Feature Enhancement}  \label{R3}
We further enhance  $\mathcal{F}_f$ with the FE module so that we can use the enhanced feature to construct high-quality frame. The FE module is composed of ($L+1$) convolution layers with $C$ filters whose sizes are all $3\times3$ and followed by a LeakyReLU activation function, $L$ spatial dual contextual blocks, and a convolution layer, as illustrated in Fig. \ref{fig_pipline}.

In the FE module, we design the spatial dual contextual block (SDCB) that implements the contextual module (CM)\cite{ref34} in dual paths to exploit the spatially contextual information and strengthen the contextual information interaction to generate the contextually-enhanced feature. Some dual-path-based methods were proposed for face image hallucination \cite{ref43} and image deraining \cite{ref44}. However, they either lack information interaction between paths \cite{ref43} or design complex interaction blocks \cite{ref44}, which limits their  efficiency. In our method, we construct SDCB via simple skip connection to strengthen information interaction between paths without increasing model parameters, guaranteeing high enhanccement performance with low complexity.

\begin{figure}
	\centerline{\includegraphics[width=9.0cm]{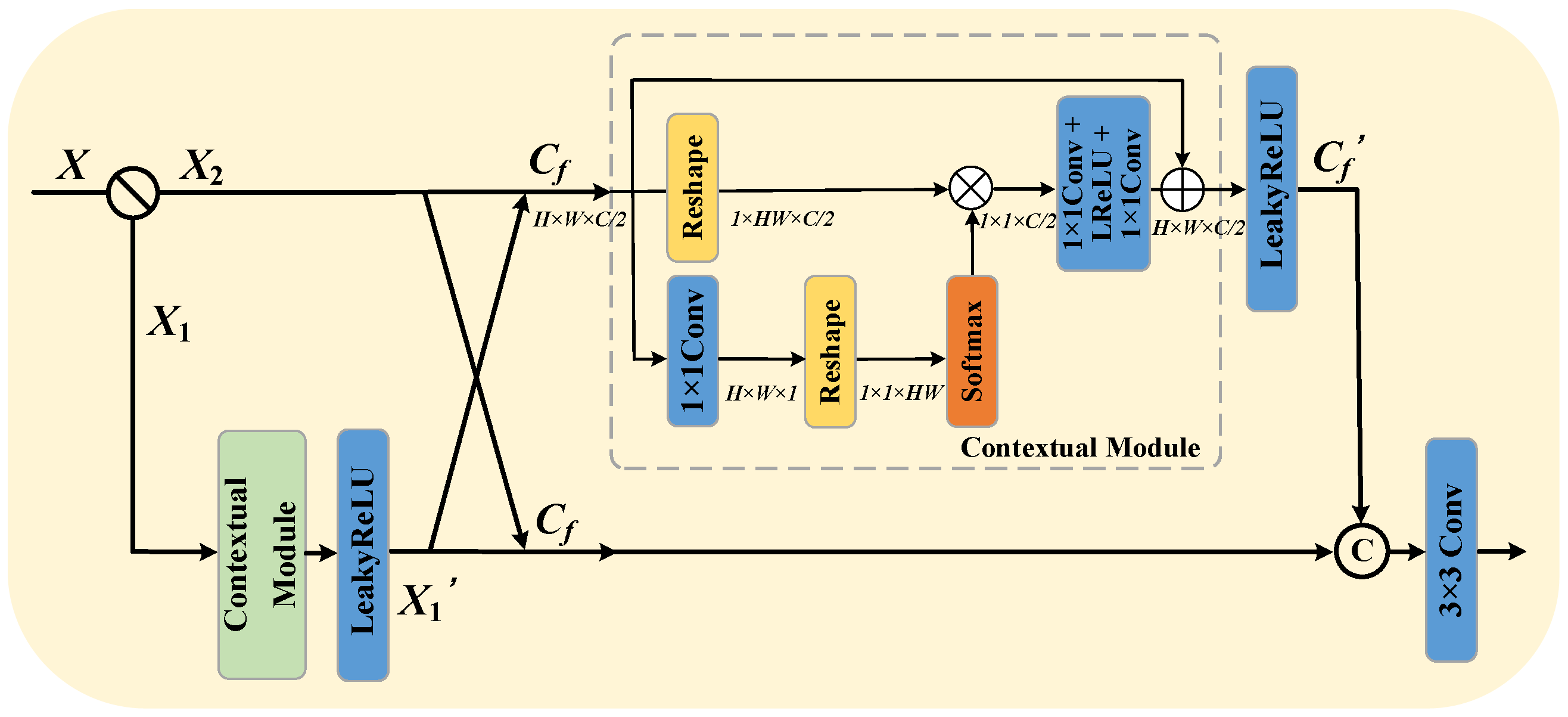}}
	\caption{Structure of SDCB.}
	\label{figRCB}
	\vspace{-1em}
\end{figure}

The architecture of SDCB is illustrated in Fig. \ref{figRCB}.  Specifically, the input feature $X$ $\in{\mathbb{R}}^{H \times W \times C}$ is split into two features $X_1$ and $X_2$ $\in{\mathbb{R}}^{H \times W \times C/2}$. Then, $X_1$ is fed into one CM followed by a LeakyReLU activation function to obtain the contextual-aware feature ${X}_{1}^{'}$. Next, ${X}_{1}^{'}$ is added to $X_2$ to obtain the feature $C_f$ $\in{\mathbb{R}}^{H \times W \times C/2}$. After that, $C_f$ is fed into another CM followed by a LeakyReLU activation function to obtain the contextual-aware feature ${C'_f}$. Finally, ${C'_f}$ and $C_f$ are fused via the concatenation and a convolution layer with $C$ filters whose size is $3\times3$ to generate the enhanced feature.

\begin{table*}[t]
	\vspace{-1.5em}
	\caption{Comparison results in terms of $\Delta$PSNR (dB) and $\Delta$SSIM ($\times$$10^{-4}$)} \label{Compare}
	\setlength{\tabcolsep}{1.4mm}
	\renewcommand\arraystretch{.85}
	\fontsize{7}{9}\selectfont
	\centering
	\vspace{-1.0em}
		\begin{tabular}{cccccccccccccccccccccccccc}
			\toprule[0.2mm]
			\toprule[0.2mm]
			\multirow{2}{*}{\textbf{QP}} &\multirow{2}{*}{\textbf{Class}}   & \multirow{2}{*}{\textbf{Sequence}}    & \multicolumn{2}{c}{\textbf { DCAD }}  & \multicolumn{2}{c}{\textbf { MFQE }} & \multicolumn{2}{c}{\textbf { MFQE2.0 }} & \multicolumn{2}{c}{\textbf { STDF-R3L }} & \multicolumn{2}{c}{\textbf { RFDA }}  &\multicolumn{2}{c}{\textbf { CF-STIF }}  &\multicolumn{2}{c}{\textbf { Ours }}  \\
			\cmidrule(r){4-5}   \cmidrule(r){6-7}    \cmidrule(r){8-9}  \cmidrule(r){10-11}  \cmidrule(r){12-13}  \cmidrule(r){14-15}  \cmidrule(r){16-17}  \cmidrule(r){18-19}
			&  &  & PSNR & SSIM  & PSNR & SSIM  & PSNR & SSIM    & PSNR & SSIM  & PSNR & SSIM  & PSNR & SSIM    & PSNR & SSIM   \\
			\hline
			\multirow{19}{*}{37}
			& \multirow{2}{2cm}{\centering A \\ (2560 $\times$ 1600)}
			& { Traffic } &  0.31 & 67 & 0.50 & 90  & 0.59 & 102 & 0.74  & 122 & 0.80 & 128 & 0.78 & 126   & 0.84 & 132  \\
			\cline{3-19}  & & {  PeopleOnStreet} & 0.50 & 95 & 0.80 & 137  & 0.92 & 157 & 1.25 & 202 & 1.44 & 222 & 1.40 & 216      & 1.53 & 233   \\
			\cline{2-19}  & & { Kimono } & 0.28 & 78 & 0.50 & 113 & 0.55 & 118 & 0.91 & 173 & 1.02 & 186 & 0.99 & 181   & 1.06 & 194   \\
			\cline{3-19}
			& \multirow{3}{2cm}{\centering B \\ (1920 $\times$ 1080)}   & { ParkScene } & 0.16 & 50 & 0.39 & 103 & 0.46 & 123 & 0.61 & 145 & 0.64 & 158 & 0.65 & 160   & 0.71 & 179   \\
			\cline{3-19} &  & { Cactus } & 0.26 & 58 & 0.44 & 88  & 0.50 & 100 & 0.77 & 142 & 0.83 & 149 & 0.83 & 150    & 0.82 & 152   \\
			\cline{3-19} & & { BQTerrace } & 0.28 & 50 & 0.27 & 48 & 0.40 & 67 & 0.63 & 105 & 0.65 & 106 & 0.71 & 117   & 0.68 & 116   \\
			\cline{3-19} & & { BasketballDrive } & 0.31 & 68 & 0.41 & 80  & 0.47 & 83 & 0.80 & 134 & 0.87 & 140 & 0.88 & 143    & 0.92 & 150   \\
			\cline{2-19} & & { RaceHorses } & 0.28 & 65 & 0.34 & 55  & 0.39 & 80 & 0.52 & 116 & 0.48 & 123 & 0.59 & 154   & 0.64 & 164   \\
			\cline{3-19}
			& \multirow{2}{2cm}{\centering C \\ (832 $\times$ 480)}  & { BQMall } &  0.34 & 88 & 0.51 & 103 & 0.62 & 120 & 0.90 & 186 & 1.09 & 197 & 1.11 & 197    & 1.13 & 207   \\
			\cline{3-19}  & & { PartyScene } & 0.16 & 48 & 0.22 & 73  & 0.36 & 118 & 0.67 & 192 & 0.66 & 188  & 0.75 & 210     & 0.81 & 229  \\
			\cline{3-19}  & & { BasketballDrill } & 0.39 & 78 & 0.48 & 90 & 0.58 & 120 & 0.79 & 146 & 0.88 & 167 & 0.94 & 173   & 0.89 & 171   \\
			\cline{2-19}  & & { RaceHorses } & 0.34 & 83 & 0.51 & 113  & 0.59 & 143 & 0.81 & 194 & 0.85 & 211 & 0.91 & 227   & 0.97 & 248  \\
			\cline{3-19}
			& \multirow{2}{2cm}{\centering D \\ (416 $\times$ 240)}   & { BQSquare } & 0.20 & 38 & -0.01 & 15 & 0.34 & 65 & 0.80 & 121 & 1.05 & 139 & 1.05 & 142  & 1.22 & 159    \\
			\cline{3-19}  & & { BlowingBubbles } & 0.22 & 65 & 0.39 & 120  & 0.53 & 170 & 0.74 & 224 & 0.78 & 240 & 0.78 & 238    & 0.83 & 263   \\
			\cline{3-19}  & & { BasketballPass } & 0.35 & 85 & 0.63 & 138  & 0.73 & 155 & 1.05 & 211 & 1.12 & 223 & 1.20 & 239   & 1.23 & 251   \\
			\cline{2-19}  & & { FourPeople } & 0.51 & 78 & 0.66 & 85  & 0.73 & 95 & 1.00 & 129 & 1.13 & 136 & 1.04 & 127  & 1.02 & 130   \\
			\cline{3-19}
			& \multirow{2}{2cm}{\centering E \\ (1280 $\times$ 720)}  & { Johnny } & 0.41 & 50 & 0.55 & 55 & 0.60 & 68 & 0.88 & 102 & 0.90 & 94 & 0.88 & 97    & 0.83 & 89   \\
			\cline{3-19}  & & { KristenAndSara } & 0.52 & 70 & 0.66 & 75  & 0.75 & 85 & 1.05 & 111 & 1.19 & 115 & 1.10 & 108    & 1.11 & 113    \\
			\cline{2-19} & & \textbf{ Average } & 0.32 & 67 & 0.46 & 88  & 0.56 & 109 & 0.83 & 153 & 0.91 & 162 & 0.92 & 167    & \textbf{0.96} & \textbf{177}   \\
			\cline{1-19}
			\multirow{1}{*}{42} &  & \textbf{ Average } & 0.32 & 109 & 0.44 & 130 & 0.59 & 165 & 0.74 & 199 & 0.82 & 220 & 0.89 & 232   & \textbf{0.92} & \textbf{236}   \\
			\cline{1-19}
			\multirow{1}{*}{32} &  & \textbf{ Average } & 0.32 & 44 & 0.43 & 58  & 0.52 & 68 & 0.86 & 107  & 0.87 & 107  & 0.95 & 118    & \textbf{1.00} & \textbf{126}  \\
			\cline{1-19}
			\multirow{1}{*}{27}& & \textbf{ Average } & 0.32 & 30 & 0.40 & 34  & 0.49 & 42 & 0.77 & 64 & 0.82 & 68 & 0.94 & 78     &  \textbf{0.99} &  \textbf{83}  \\
			\cline{1-19}
			\multirow{1}{*}{22}& & \textbf{ Average }  & 0.31 & 19 & 0.31 & 19  & 0.46 & 27 & 0.65 & 34 & 0.76 & 42 & 0.85 & 46    &  \textbf{0.90} &  \textbf{50}   \\
			\toprule[0.2mm]
			\toprule[0.2mm]
		\end{tabular}
		\vspace{-1.5em}
\end{table*}
\vspace{-.55em}
\section{Experimental Results}

\subsubsection{Experiment Setup}

Similar to \cite{ref24, ref18, ref21-1}, we conducted our experiments on MFQE2.0 dataset \cite{ref24} that consists of 108 training videos and 18 testing videos of the Joint Collaborative
Team on Video Coding \cite{ref28-1}. All the videos are compressed by HM16.5 \cite{ref28} under the LDP configuration. To evaluate the performance for different coding scenarios, the compression is carried out with 5 quantization parameters (QPs), i.e., 22, 27, 32, 37, and 42. Both the raw and compressed videos are cropped into 128$\times$128 patches as the training pairs. We adopt flip and rotation as data augmentation techniques to expand the training dataset as \cite{ref18} did. In addition, two quantitative metrics PSNR and SSIM are used to evaluate the distortion on the luminance component as the previous work \cite{ref24, ref18,ref21-1} did.
For the network settings, 7 successive frames are used to construct the input sequence. The number of SDCB in the FE module is 3 and the channel number of feature is 64.

\subsubsection{Model Training}

We use the Charbonnier loss \cite{ref45} to train our model. Additionally, the batch size is set to 32 and the learning rate is set to $1\times 10^{-4}$. Our model is trained by Adam optimizer \cite{ref29} with the exponential decay rates $\beta_1=0.9$ and $\beta_2=0.999$, {the compensation factor} $\varepsilon = 1\times 10^{-8}$, and the total number of iterations is $3\times 10^5$. The proposed models are developed based on Pytorch with two NVIDIA GeForce RTX 3090 GPUs and Intel(R) Xeon(R) Gold 6133 CPU.

%
%

\subsubsection{Comparison to State-of-the-arts}

\begin{figure}[t]
	\centering
	\setlength{\abovecaptionskip}{-0.05cm}
	\begin{minipage}[b]{0.235\linewidth}
		\centering
		\centerline{\epsfig{figure=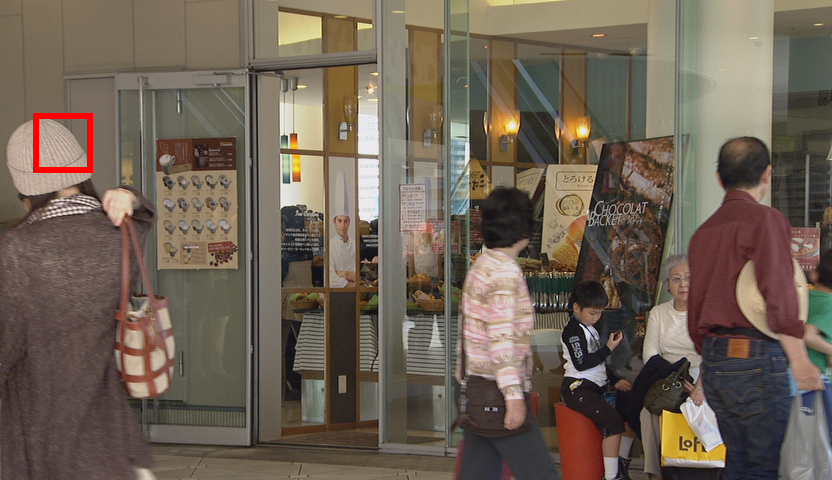, width=2.08cm}}
		\scriptsize{BQMall}
	\end{minipage}
	\begin{minipage}[b]{0.135\linewidth}
		\centering
		\centerline{\epsfig{figure=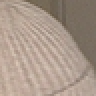,width=1.18cm}}
		\scriptsize{Raw}
	\end{minipage}
	\begin{minipage}[b]{0.135\linewidth}
		\centering
		\centerline{\epsfig{figure=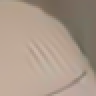,width=1.18cm}}
		\scriptsize{STDF-R3L}
	\end{minipage}
	\begin{minipage}[b]{0.135\linewidth}
		\centering
		\centerline{\epsfig{figure=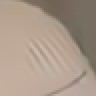,width=1.18cm}}
		\scriptsize{RFDA}
	\end{minipage}
	\begin{minipage}[b]{0.135\linewidth}
		\centering
		\centerline{\epsfig{figure=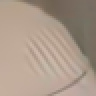,width=1.18cm}}
		\scriptsize{CF-STIF}
	\end{minipage}
	\begin{minipage}[b]{0.135\linewidth}
		\centering
		\centerline{\epsfig{figure=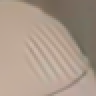,width=1.18cm}}
		\scriptsize{Ours}
	\end{minipage}
		
	\begin{minipage}[b]{0.235\linewidth}
		\centering
		\centerline{\epsfig{figure=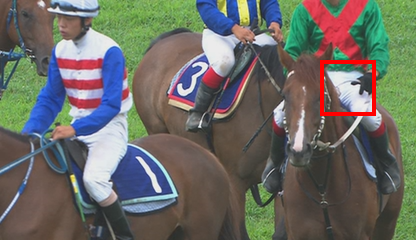, width=2.08cm}}
		\scriptsize{RaceHorses}
	\end{minipage}
	\begin{minipage}[b]{0.135\linewidth}
		\centering
		\centerline{\epsfig{figure=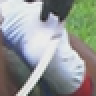,width=1.18cm}}
		\scriptsize{Raw}
	\end{minipage}
	\begin{minipage}[b]{0.135\linewidth}
		\centering
		\centerline{\epsfig{figure=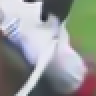,width=1.18cm}}
		\scriptsize{STDF-R3L}
	\end{minipage}
	\begin{minipage}[b]{0.135\linewidth}
		\centering
		\centerline{\epsfig{figure=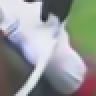,width=1.18cm}}
		\scriptsize{RFDA}
	\end{minipage}
	\begin{minipage}[b]{0.135\linewidth}
		\centering
		\centerline{\epsfig{figure=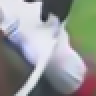,width=1.18cm}}
		\scriptsize{CF-STIF}
	\end{minipage}
	\begin{minipage}[b]{0.135\linewidth}
		\centering
		\centerline{\epsfig{figure=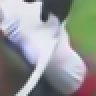,width=1.18cm}}
		\scriptsize{Ours}
	\end{minipage}
	
	\begin{minipage}[b]{0.235\linewidth}
		\centering
		\centerline{\epsfig{figure=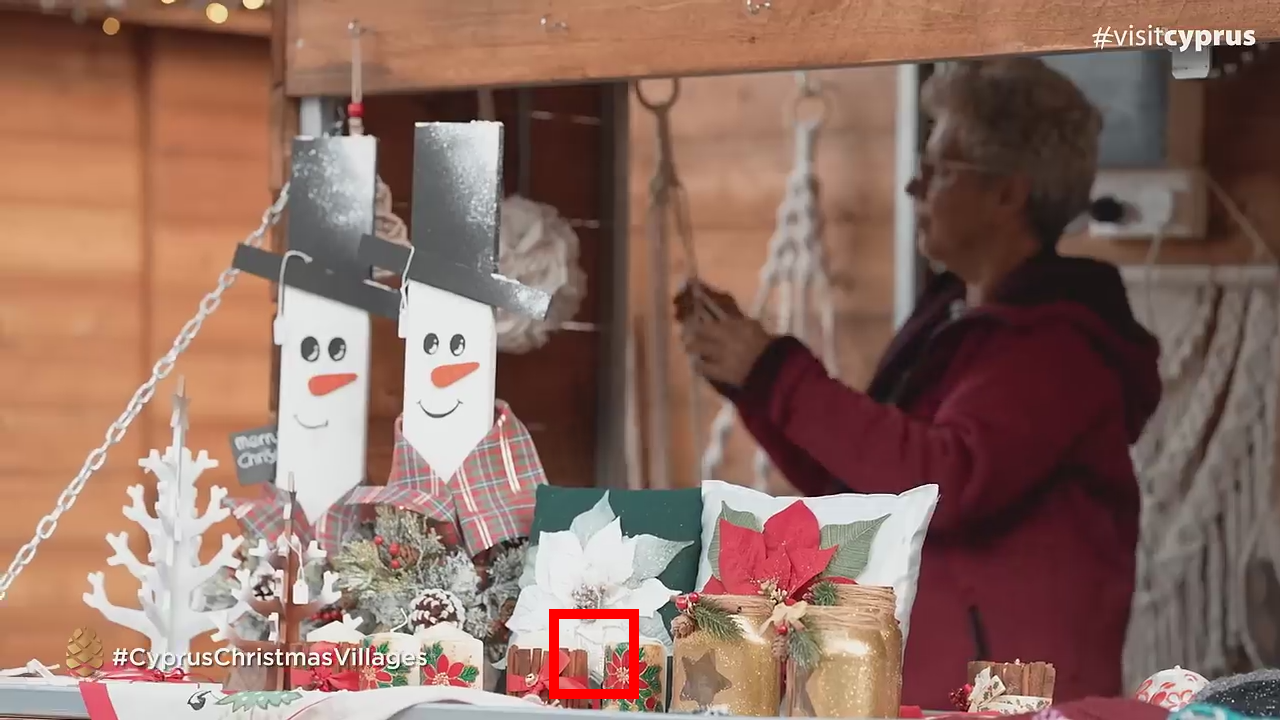, width=2.08cm}}
		\scriptsize{Real-world video-1}  
	\end{minipage}
	\begin{minipage}[b]{0.135\linewidth}
		\centering
		\centerline{\epsfig{figure=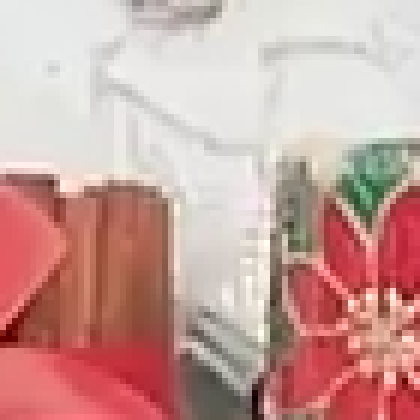,width=1.18cm}}
		\scriptsize{LQ}  
	\end{minipage}
	\begin{minipage}[b]{0.135\linewidth}
		\centering
		\centerline{\epsfig{figure=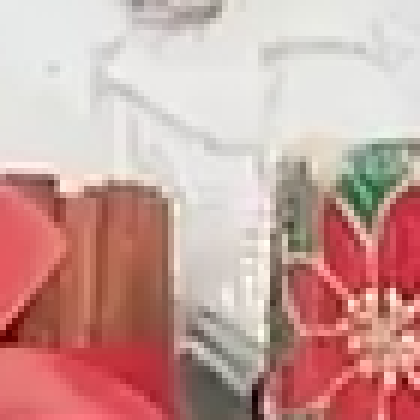,width=1.18cm}}
		\scriptsize{STDF-R3L}
	\end{minipage}
	\begin{minipage}[b]{0.135\linewidth}
		\centering
		\centerline{\epsfig{figure=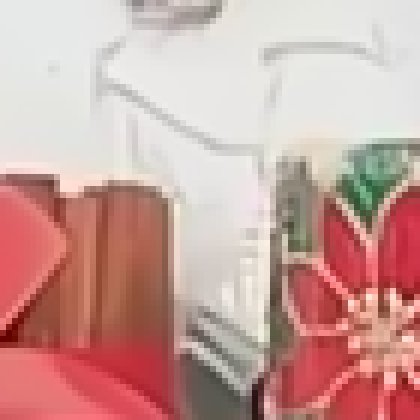,width=1.18cm}}
		\scriptsize{RFDA}
	\end{minipage}
	\begin{minipage}[b]{0.135\linewidth}
		\centering
		\centerline{\epsfig{figure=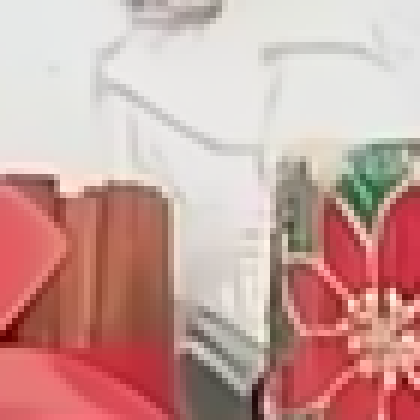,width=1.18cm}}
		\scriptsize{CF-STIF}
	\end{minipage}
	\begin{minipage}[b]{0.135\linewidth}
		\centering
		\centerline{\epsfig{figure=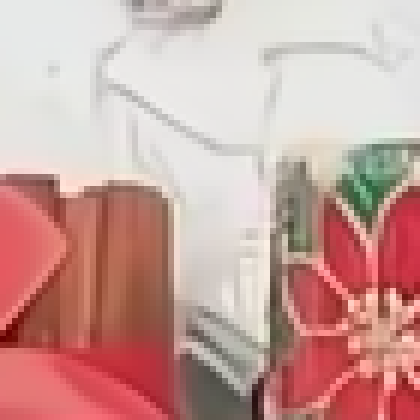,width=1.18cm}}
		\scriptsize{Ours}
	\end{minipage}
	
	\begin{minipage}[b]{0.235\linewidth}
		\centering
		\centerline{\epsfig{figure=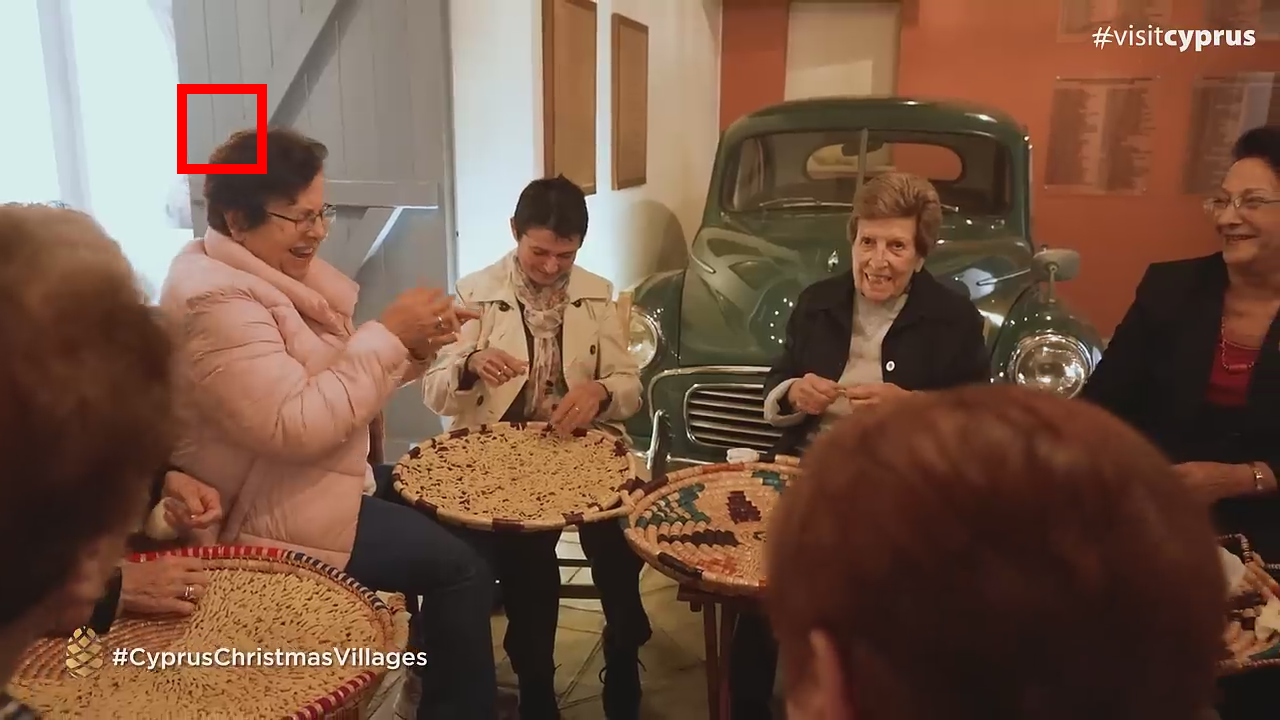, width=2.08cm}}
		\scriptsize{Real-world video-2}  
	\end{minipage}
	\begin{minipage}[b]{0.135\linewidth}
		\centering
		\centerline{\epsfig{figure=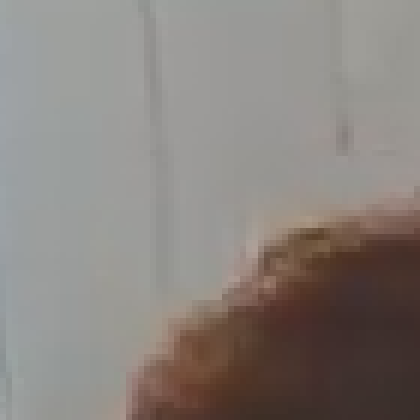,width=1.18cm}}
		\scriptsize{LQ}  
	\end{minipage}
	\begin{minipage}[b]{0.135\linewidth}
		\centering
		\centerline{\epsfig{figure=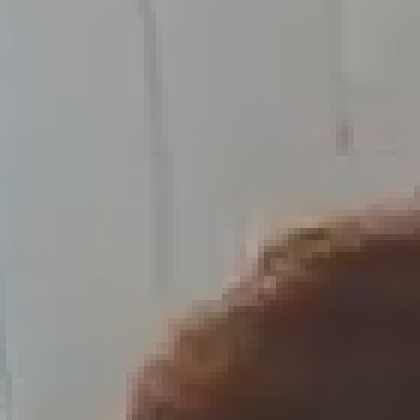,width=1.18cm}}
		\scriptsize{STDF-R3L}
	\end{minipage}
	\begin{minipage}[b]{0.135\linewidth}
		\centering
		\centerline{\epsfig{figure=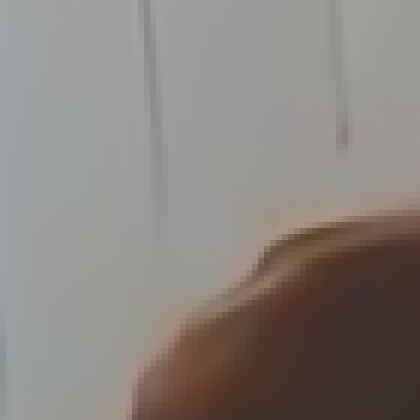,width=1.18cm}}
		\scriptsize{RFDA}
	\end{minipage}
	\begin{minipage}[b]{0.135\linewidth}
		\centering
		\centerline{\epsfig{figure=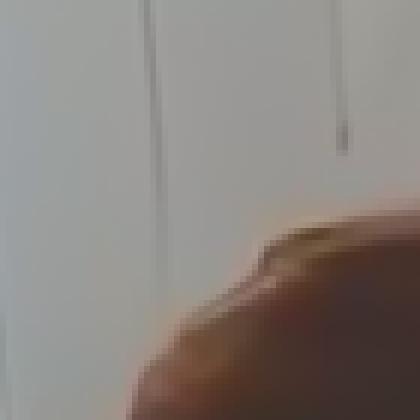,width=1.18cm}}
		\scriptsize{CF-STIF}
	\end{minipage}
	\begin{minipage}[b]{0.135\linewidth}
		\centering
		\centerline{\epsfig{figure=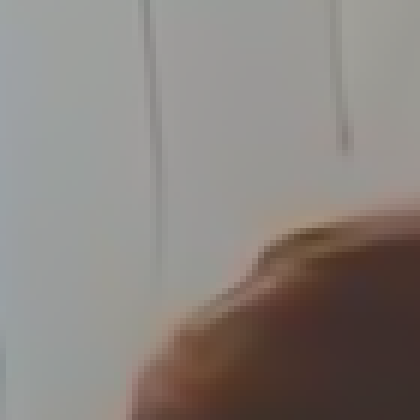,width=1.18cm}}
		\scriptsize{Ours}
	\end{minipage}
	
	\caption{Comparison of visual results ($QP$=37). The $1^{st}$ row : The $193^{th}$ frame of $BQMall$ video. The $2^{nd}$ row: The $19^{th}$ frame of $RaceHorses$ video. The $3^{rd}$ row: The $168^{th}$ frame of real-world compressed video-1. The $4^{th}$ rows: The $1,082^{th}$ frame of real-world compressed video-2.}
	\label{fig_vision}
	\vspace{-0.5em}
\end{figure}

We compare our proposed method with one single-frame-based method, i.e., deep CNN-based auto decoder (DCAD) \cite{ref11}, and some multi-frame-based methods, i.e., MFQE \cite{ref14}, MFQE2.0 \cite{ref24}, STDF-R3L\cite{ref18}, RFDA\cite{ref21-1}, and coarse-to-fine spatio-temporal information fusion (CF-STIF)\cite{ref32}. The results of $\Delta$PSNR and $\Delta$SSIM are presented in Table \ref{Compare}, where the results of those compared approaches are offered by \cite{ref21-1,ref32}. It can be seen from Table \ref{Compare} that our method offers the best performance in terms of $\Delta$PSNR and $\Delta$SSIM (on average), compared with these state-of-the-art methods on the testing videos at five QPs.  Additionally, we present some visual results in Fig. \ref{fig_vision} on testing compressed videos and real-world compressed videos to make a comprehensive comparison. The real-world videos were downloaded from YouTube with 1280$\times$720 resolution. These results are obtained by using our proposed method and three competitive methods, i.e., STDF-R3L, RFDA and CF-STIF. It can be observed from Fig. \ref{fig_vision} that compared with the other three methods, our method reduces more compression artifacts and achieves better visual results.

\begin{table}[t]
\vspace{-1em}
\caption{Ablation study at $QP$=37 with STDF as baseline} \label{Ablation}
\setlength{\tabcolsep}{1.7mm}
\renewcommand\arraystretch{1.0}
\fontsize{7}{9}\selectfont
\centering
\vspace{-1.2em}
\begin{tabular}{cc|cccc}
	\toprule[0.2mm]
	\bottomrule[0.2mm] 
	  IntraGFA+InterGFF  &  FE    &$\Delta$PSNR (dB)   & $\Delta$SSIM ($\times10^{-4}$) \\
	\hline 
	  - & -  & 0.77 & 139   \\
	  \checkmark & -  & 0.88 & 160   \\
	   \checkmark & \checkmark &  \textbf{0.96}  & \textbf{177}  \\
	\toprule[0.2mm]
	\bottomrule[0.2mm]
\end{tabular}
\vspace{-2em}
\end{table}

\begin{table}[t]
	\vspace{-0.5em}
	\caption{Verification of using different numbers of SDCB at $QP$=37} \label{Abl_Com}
	\setlength{\tabcolsep}{2.0mm}
	\renewcommand\arraystretch{1.0}
	\fontsize{7}{9}\selectfont
	\centering
	\vspace{-1.0em}
	\begin{tabular}{cccccccccccccccccccccccccc}
		\toprule[0.2mm]
		\bottomrule[0.2mm]
		&   &$\Delta$PSNR (dB)   & $\Delta$SSIM ($\times10^{-4}$)   & Param. (K) \\
		\hline
		& Ours-1 &  0.92  & 169 & 1251  \\
		& Ours-2 &  0.94  & 173 & 1327  \\
		& Ours &  \textbf{0.96}  & \textbf{177} & 1403  \\
		\toprule[0.2mm]
		\bottomrule[0.2mm]
	\end{tabular}
\vspace{-1em}
\end{table}

\begin{table}[t]
	\vspace{-0.5em}
	\caption{Comparison of parameters, FLOPs and inference speed (fps)} \label{Abl_FPS}
	\setlength{\tabcolsep}{2.3mm}
	\renewcommand\arraystretch{1.0}
	\fontsize{7}{9}\selectfont
	\centering
	\vspace{-1.2em}
	\begin{tabular}{ccccccccccc}
		\toprule[0.2mm]
		\bottomrule[0.2mm]
		&  {\multirow{2}{*} {{Param.}(K)}}  &{\multirow{2}{*} {{FLOPs}(G)}}  & \multicolumn{3}{c}{{FPS} @Different Resolution} \\
		\cline { 4 - 6 } & & & Class C ${\ }$   & Class D  & Class E  \\
		\hline
		STDF-R3L & 1275 & 83.5  & 17.37 & 37.03  & 4.24 \\
		RFDA & 1270 & 56.8  & 15.41 & 32.14 & 3.74 \\
		CF-STIF & 2200 & 194.6 & 9.16 & 26.32 & 2.05 \\
		Ours & 1403 & 143.2  & 11.00 & 28.54 & 2.43 \\ 
		\toprule[0.2mm]
		\bottomrule[0.2mm]
	\end{tabular}
	\vspace{-1.5em}
\end{table}

\subsubsection{Ablation Study}
We conduct the ablation study to validate the effectiveness of the IntraGFA, InterGFF and FE modules over all the test videos. In this experiment, we use the STDF module of STDF-R3L\cite{ref18} as the baseline to enhance the compressed videos. The results of baseline method are presented in the first row in Table \ref{Ablation}. Because the IntraGFA and InterGFF modules can not separately work in our model, we combine them together to verify the effectiveness. The corresponding results are given in Table \ref{Ablation}. It is found from Table \ref{Ablation} that IntraGFA and InterGFF modules obtain 0.11dB $\Delta$PSNR gain (on average) over the baseline, which demonstrates their effectiveness. Besides, we introduce our FE module with three SDCBs to the combined IntraGFA and InterGFF modules to construct our proposed model. It is found from Table \ref{Ablation} that introducing the FE module obtains further improves the performance, demonstrating its effectiveness.

In addition, we verify the effectiveness of the number of designed SDCB.  Specifically, we use one SDCB and two SDCBs to construct two corresponding models, denoted as \emph{Ours-1} and \emph{Ours-2}, respectively.
The corresponding results are given in Table \ref{Abl_Com}. It is found from Table \ref{Abl_Com} that the method \emph{Ours-1} i.e., using one SDCB, performs better than RFDA and CF-STIF, as proved by the achieved higher performance gain. In addition, the method \emph{Ours-2}, i.e., using two SDCBs, performs better than \emph{Ours-1}, which indicates that using more SDCBs can effectively improve the enhancement performance.


\subsubsection{Complexity Comparison}
We provide model parameters, FLOPs, and inference speed of our method and three comparable methods, i.e., STDF-R3L, RFDA and CF-STIF, in Table \ref{Abl_FPS}, where all the compared methods adopt 7 successive frames to generate enhanced video. The FLOPs are calculated on the Class-D video and the inference speed is evaluated in terms of the frame per second (FPS). It is found from Table \ref{Abl_FPS} that our method achieves fewer parameters and FLOPs, but higher inference speed than CF-STIF. Although our method is more complicated than STDF-R3L and RFDA, it is still competitive for practical applications due to the higher enhancement performance.

\vspace{-0.8em}
\section{Conclusion}

In this paper, we proposed temporal group alignment and fusion network to enhance the quality of compressed video. With our proposed model, the long-short term correlations between frames are explored to generate effective features to compose high-quality videos. The complexity of our method accordingly increases due to extracting these features, which will limit the application efficiency of our method. We will focus on the design of low-complexity modules for our enhancement model in the future work.

\clearpage

\end{document}